\documentclass[conference]{IEEEtran}
\IEEEoverridecommandlockouts

\usepackage{cite}
\usepackage{amsmath,amssymb,amsfonts}
\usepackage{algorithmic}
\usepackage{graphicx}
\graphicspath{{}{IEEE-conference-template-062824/}}
\usepackage{textcomp}
\usepackage{xcolor}
\usepackage{booktabs}
\usepackage{multirow}
\usepackage{float}
\usepackage{url}
\usepackage{comment}

\def\BibTeX{{\rm B\kern-.05em{\sc i\kern-.025em b}\kern-.08em
    T\kern-.1667em\lower.7ex\hbox{E}\kern-.125emX}}

\begin{document}

\title{MethaneFuse: Learning from Multi-Sensor Satellite Observations for Methane Plume Detection}

\author{
\IEEEauthorblockN{Yuyao Wang}
\IEEEauthorblockA{\textit{Department of Electrical and} \\
\textit{Computer Engineering} \\
\textit{University of Alberta}\\
Edmonton, Canada \\
yuyao16@ualberta.ca}
\and
\IEEEauthorblockN{Juliana Y. Leung}
\IEEEauthorblockA{\textit{Department of Civil and} \\
\textit{Environmental Engineering} \\
\textit{University of Alberta}\\
Edmonton, Canada \\
juliana2@ualberta.ca}
\and
\IEEEauthorblockN{Di Niu}
\IEEEauthorblockA{\textit{Department of Electrical and} \\
\textit{Computer Engineering} \\
\textit{University of Alberta}\\
Edmonton, Canada \\
dniu@ualberta.ca}
}

\IEEEaftertitletext{%
\begin{center}
\footnotesize
\copyright~2026 IEEE. Personal use of this material is permitted.
Permission from IEEE must be obtained for all other uses, in any current or future media,
including reprinting/republishing this material for advertising or promotional purposes,
creating new collective works, for resale or redistribution to servers or lists,
or reuse of any copyrighted component of this work in other works.
\end{center}
}

\maketitle

\begin{abstract}
Methane plume detection from satellite imagery is fundamentally constrained by incomplete observations: although public satellites provide complementary spatial, spectral, and atmospheric evidence, real plume cases rarely contain fully paired multi-sensor measurements because of revisit schedules, cloud coverage, acquisition quality, and the transient nature of emissions. Most learning-based detectors are developed around single-sensor inputs, especially Sentinel-2 (S2), but many reported plume cases do not have valid S2 observations. This creates a challenging partial-observation learning problem where different plume cases are captured by different subsets of heterogeneous sensors rather than by complete multi-sensor stacks.

To address this problem, we construct \textbf{MethaneUnion}, a temporal multi-sensor dataset built from Carbon Mapper plume reports and matched S2, Landsat 8/9 (L8/9), EMIT, and Sentinel-5P (S5P) observations. Built on MethaneUnion, we propose \textbf{MethaneFuse}, a learning framework for methane plume detection from heterogeneous satellite observations where fully paired multi-sensor measurements are almost never available.

MethaneUnion expands usable coverage from 3,211 valid S2-matched plume cases to 8,981 reported plume cases with multi-sensor observations. Experiments show that MethaneFuse consistently outperforms independently trained sensor predictors, heuristic fusion strategies, and generic Earth observation representation transfer. At the representative 480~m setting, MethaneFuse achieves 84.87 F1 and 93.62 AUROC, improving over the strongest baseline by 5.65 F1 and 8.30 AUROC points while reducing false positives by 8.19 points. Sensor-availability experiments further show that MethaneFuse improves detection when S2 is available and transfers plume knowledge to L8/9, EMIT, and S5P when S2 is unavailable. These results demonstrate that learning from incomplete heterogeneous sensor observations provides a practical and effective alternative to conventional single-sensor methane detection pipelines. \textit{The dataset and code are available at \url{https://github.com/yuyao-wang/MethaneFuse}.}
\end{abstract}


\section{Introduction}

Methane is a short-lived but powerful greenhouse gas, making rapid emission reduction a high-impact path for limiting near-term warming~\cite{unep2021globalmethane}.
Reliable methane monitoring is therefore essential for locating large or intermittent emission sources, verifying reported emissions, and prioritizing infrastructure repair~\cite{iea2025tracker}.
However, detecting methane plumes from satellite observations is difficult because plume signals are transient, weak, and strongly affected by sensor resolution, spectral coverage, surface background, and acquisition conditions~\cite{varon2021s2,gorrono2023s2}.
Unlike land-cover classes or common objects in images, methane plumes are not persistent scene elements with stable shapes or clear visual boundaries.

Existing learning-based plume detectors commonly rely on Sentinel-2 (S2) because its fine-resolution multispectral shortwave infrared (SWIR) bands are useful for plume detection~\cite{vaughan2024ch4net,rouetleduc2024vitmethane,radman2023s2metnet}. However, valid S2 observations are often unavailable for transient plume reports because of revisit timing, cloud cover, acquisition geometry, and quality filtering. In the Carbon Mapper plume reports collected from 2016--2025, more than 24,000 reported plume cases yield only 3,211 valid S2-matched cases after sensor matching and quality filtering, meaning that most reported plumes cannot be used by S2-only detectors~\cite{carbonmapper,sentinel2l2a}. Other public satellites can provide additional methane-related evidence: Landsat 8/9 (L8/9) offers multispectral observations at approximately 30~m resolution, EMIT provides hyperspectral observations at approximately 60~m resolution, and Sentinel-5P (S5P) provides coarse CH$_4$ products with kilometer-scale footprints~\cite{schuit2023automated,s5pch4,bian2025workflow,bian2025analytics,ruzicka2023starcop,thorpe2023emit,joyce2023prisma}. 

The challenge is that this additional evidence is rarely available as complete multi-sensor measurements for the same plume case. Among the Carbon Mapper plume reports from 2016 to 2025, only four cases have observations from all four sensors. Therefore, the practical learning problem is to learn methane-relevant cues from plume cases observed by different satellite sources, rather than from plume cases where all sensors are available together. This motivates our central question: \textit{How can methane plume detectors learn from heterogeneous multi-sensor satellite observations when fully paired sensor measurements are almost unavailable?}

In this paper, we first construct \textbf{\textit{MethaneUnion}}, a temporal multi-sensor satellite dataset for methane plume detection. \textit{MethaneUnion} reflects the real monitoring setting in which different plume cases are captured by different subsets of public satellites. Built on \textit{MethaneUnion}, we propose \textbf{\textit{MethaneFuse}}, a learning framework that learns methane plume evidence from available sensor subsets through sensor-native representation learning
and transfers it to scale-controlled plume classification and segmentation with lightweight sensor-aware adaptation.

\noindent Our contributions are summarized as follows:
\begin{itemize}
\item We construct \textbf{MethaneUnion}, to the best of our knowledge, the first methane plume dataset specifically designed for temporal multi-sensor learning from naturally available satellite observations.
Using Carbon Mapper plume reports as methane anchors, we match S2, L8/9, EMIT, and S5P observations near reported plume locations and times, organize each sensor stream into plume-time, recent-reference, and earlier-reference views, and derive plume-positive and local non-plume background samples for supervised evaluation.
\textit{MethaneUnion} expands usable coverage from 3,211 valid S2-matched plume cases to 8,981 reported plume cases with multi-sensor observations.

\item We propose \textbf{MethaneFuse} for learning methane plume knowledge from heterogeneous multi-sensor plume observations.
During Stage~1 representation learning, each available sensor is kept in its native spatial--spectral form and encoded independently, while unavailable sensors are masked out.
A masked sensor-set fusion module learns to aggregate plume evidence across the available sensor views without requiring complete four-sensor measurements for each target location and time.

\item We introduce a lightweight \textbf{sensor-aware adaptation} stage for downstream plume detection at different evaluation scales.
The Stage~1 encoder is frozen, and only CLS-routed LoRA expert adapters and task heads are trained for query-level classification and plume segmentation.
This adaptation stage transfers the learned multi-sensor representation to evaluation crops with 120--960~m ground regions, while requiring only a small number of trainable parameters and separating classification fusion from segmentation-capable sensor fusion. 
\end{itemize}

We conduct extensive experiments on \textit{MethaneUnion} and summarize three main findings. First, \textit{MethaneFuse} outperforms independently trained sensor predictors, heuristic fusion strategies, and prior Earth observation pretrained models after supervised fine-tuning on the same multi-sensor data, including SatMAE~\cite{satmae2022}, AnySat~\cite{anysat2024}, and Panopticon~\cite{panopticon2025}. In the main 480~m evaluation setting, \textit{MethaneFuse} improves over the strongest baseline by 5.65 F1 points and 8.30 AUROC points while reducing false positives by 8.19 points. Second, \textit{MethaneFuse} improves detection both when S2 is available and when it is missing. In S2-present groups, \textit{MethaneFuse} improves F1 by 3.5 points for S2 alone and by 14.3 points for S2+S5P; in S2-absent groups, it improves all evaluated L8/9, EMIT, and S5P combinations, with gains up to 15.3 F1 points. Third, \textit{MethaneFuse} remains effective across evaluation crops from 120~m to 960~m on the ground, consistently outperforming the strongest heuristic fusion baseline at each scale. For plume segmentation, \textit{MethaneFuse}-Seg further improves fused IoU+ using only sensors that can provide meaningful plume-mask predictions.

\section{Related Work}

\subsection{Satellite Methane Detection and Benchmarks}

Satellite methane detection uses heterogeneous instruments with different
roles: coarse atmospheric products support broad CH$_4$ screening,
multispectral SWIR imagery supports finer-scale plume analysis, and
hyperspectral observations provide denser spectral evidence for retrieval and
mapping~\cite{schuit2023automated,bian2025workflow,bian2025analytics,varon2021s2,thorpe2023emit,ruzicka2023starcop}.
Operational systems often combine these sensors sequentially through a
tip-and-cue pipeline, where coarse detections guide higher-resolution
follow-up for localization, attribution, or quantification~\cite{schuit2023automated}.
This demonstrates sensor complementarity, but methane plume detection has not
been systematically studied as a multimodal learning problem over the
heterogeneous satellite observations available for each target location and
time.

Learning-based methane detection remains largely sensor-specific, with models
typically built around one dominant observation source, including
multispectral SWIR imagery, S2 transformers, TROPOMI screening, or
hyperspectral plume benchmarks~\cite{radman2023s2metnet,liu2025methanes2cm,zhao2025deeptransfer,rouetleduc2024vitmethane,schuit2023automated,ruzicka2023starcop,joyce2023prisma}.
Recent datasets begin to collect methane plume examples across sensors~\cite{methaneset2026}, but they mainly unify sensor-specific datasets rather than formulating methane plume detection as temporal multi-sensor learning. They do not explicitly represent each plume case by the available heterogeneous satellite observations at the same target location and time. 
\textit{MethaneUnion} targets this setting
by organizing reported plume cases into multi-sensor observations with sensor-dependent labels.

\subsection{EO Foundation Models and Multi-Sensor Fusion}

EO models learn transferable geospatial representations from large-scale satellite imagery using masked reconstruction, temporal or multispectral pretraining, cross-sensor alignment, and multimodal self-supervision~\cite{satmae2022,skysense2024}.
Recent multi-sensor and any-sensor models further support heterogeneous resolutions, spectral configurations, sensor metadata, and multimodal inputs within unified backbones~\cite{msgfm2024,anysat2024,panopticon2025}.
These models provide strong generic representations for remote-sensing tasks, especially when target semantics are persistent across sensors and acquisition conditions.

Remote-sensing fusion and missing-modality learning have also been widely studied through early, intermediate, and late fusion, cross-modal attention, hyperspectral--multispectral fusion, SAR--optical fusion, spatiotemporal fusion, modality dropout, hetero-modal embeddings, and robust multimodal transformers~\cite{ma2022missing,chen2024incompletefusion,dpmamba2025}.
These methods are highly relevant to incomplete sensor availability, but methane monitoring introduces a different target structure.
Plume evidence is transient, diffuse, and observation-conditioned: useful cues may appear as SWIR absorption, hyperspectral gas-sensitive structure, weak multispectral anomaly, or coarse atmospheric enhancement depending on the sensor and scale.
Therefore, methane evidence is not simply the common semantic intersection shared by all sensors.
\textit{MethaneFuse} addresses this gap through methane-supervised representation learning, sensor-native tokenization, and masked sensor-set fusion, allowing the model to preserve
and aggregate the union of available methane-relevant cues.

\subsection{Parameter-Efficient and Sensor-Aware Adaptation}

Parameter-efficient fine-tuning has become a practical strategy for adapting large pretrained models with limited labeled data.
LoRA freezes the backbone and learns low-rank residual updates~\cite{lora2022}, while recent geospatial studies show that LoRA, adapters, and related modules can adapt EO backbones for classification, segmentation, and change detection with lower training cost than full fine-tuning~\cite{peftgeofm2025}.
Mixture-of-experts methods provide conditional specialization by routing inputs to different experts~\cite{shazeer2017moe}, and recent parameter-efficient adaptation studies have explored mixtures of LoRA experts and layer-wise expert allocation for task-, domain-, or input-conditioned specialization~\cite{mole2024,mola2025}.
Related remote-sensing work has explored expert-based models for heterogeneous sensors, missing modalities, and multimodal foundation-model pretraining~\cite{ringmoe2025,mamol2026}.

For methane monitoring, adaptation is not only a parameter-efficiency issue.
False positives and plume cues are strongly sensor-dependent because clouds, surface materials, spectral noise, retrieval artifacts, and spatial resolution affect each satellite differently.
This motivates sensor-aware adaptation instead of applying a single shared fine-tuning module to all observations.
In \textit{MethaneFuse}, the shared methane representation is frozen during downstream adaptation, and lightweight sensor-aware LoRA expert adapters are inserted to suppress sensor-specific background artifacts while preserving cross-sensor transfer for classification and segmentation.

\section{Methodology}
\label{sec:methodology}

\begin{figure*}[t]
    \centering
    \includegraphics[width=\textwidth]{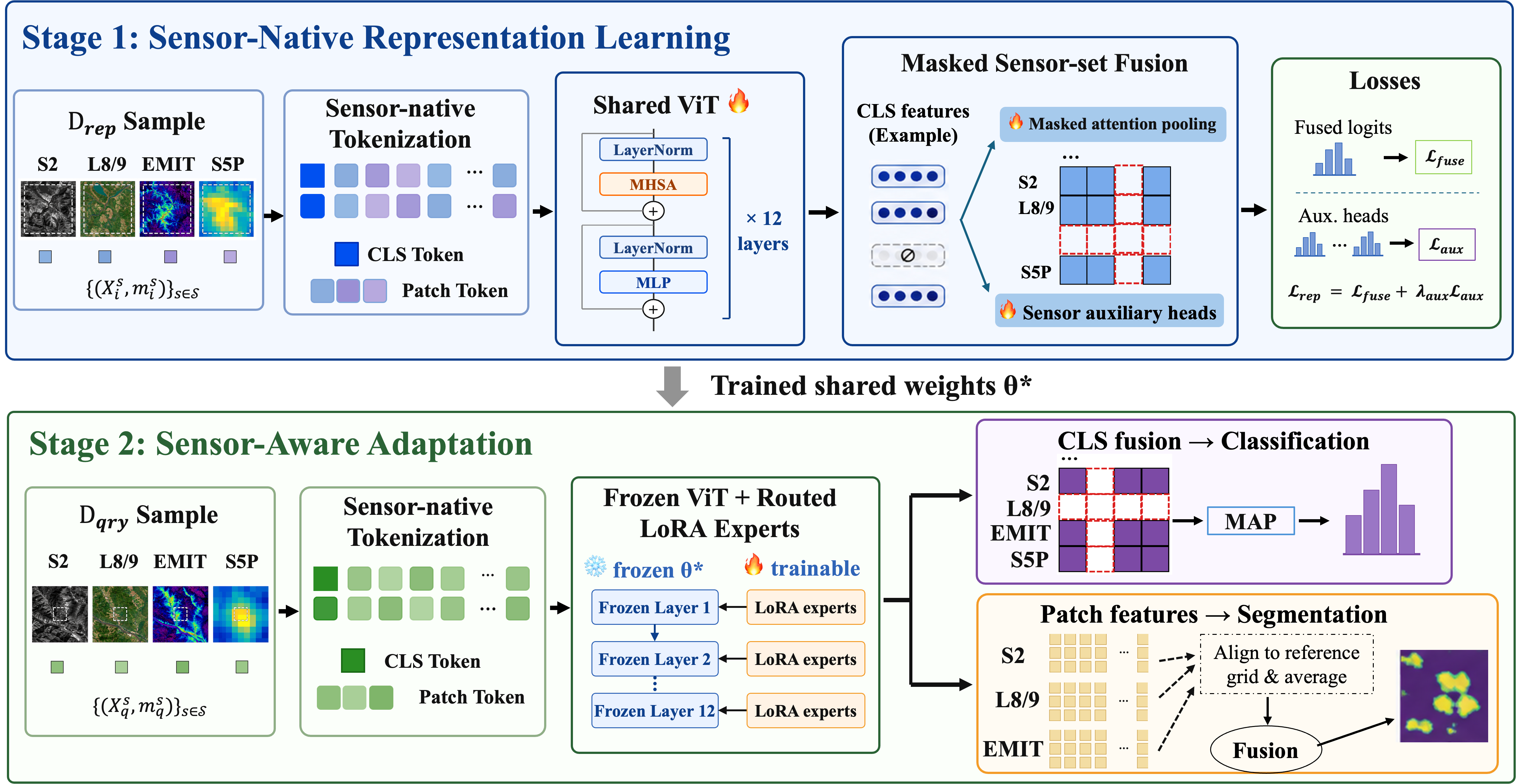}%
    \caption{
Overview of \textit{MethaneFuse}.
Stage~1 performs sensor-native representation learning on $\mathcal{D}_{\mathrm{rep}}$,
where available sensor observations are tokenized, encoded by a shared ViT backbone,
and aggregated through masked sensor-set fusion.
Stage~2 performs sensor-aware adaptation on $\mathcal{D}_{\mathrm{qry}}$,
where the Stage~1 encoder is frozen and CLS-routed LoRA expert adapters provide
lightweight input-conditioned adaptation.
Classification uses learned masked fusion over available sensor CLS features,
while segmentation uses available-sensor late fusion over aligned probability maps.
In Stage~1, $i$ indexes a training sample derived from an event, and in Stage~2,
$q$ indexes a scale-controlled query derived from an event.
}
    \label{fig:model-architecture}
\end{figure*}

We propose \textbf{\textit{MethaneFuse}}, a two-stage framework for learning from event-centered multi-sensor observations.
For each plume event or query anchor, only a naturally available subset of heterogeneous satellite sensors may be observed.
To instantiate this setting, we construct \textbf{\textit{MethaneUnion}}, an event-centered partial-observation collection built from Carbon Mapper plume reports and naturally available S2, L8/9, EMIT, and S5P observations.
\textit{MethaneUnion} is not a fully paired multi-sensor dataset; instead, it preserves missing-sensor patterns caused by revisit timing, mission coverage, clouds, acquisition quality, and the transient lifetime of methane plumes.

\textit{MethaneFuse} uses two training views derived from the same event-centered pool.
Stage~1 learns methane representations from fixed-pixel sensor-native samples, while Stage~2 adapts them to query-level classification and segmentation.

Across both stages, the central inference problem remains unchanged: predicting methane occurrence from the naturally available subset of heterogeneous sensors for each queried event. Here, a query denotes a scale-controlled prediction unit derived from an event, specified by a timestamp, center location, and geographic footprint.

\subsection{\textit{MethaneUnion} Dataset Construction}

Let $\mathcal{S}=\{\mathrm{S2}, \mathrm{L8/9}, \mathrm{EMIT}, \mathrm{S5P}\}$ denote the supported sensor set, corresponding to S2, L8/9, EMIT, and S5P. 
For each Carbon Mapper plume report, we use its timestamp, geolocation, and plume mask as the event anchor~\cite{carbonmapper}. 
Around each anchor, we collect naturally available observations from S2 Level-2A surface reflectance, L8/9 Collection 2 Level-2 surface reflectance, EMIT Level-2A hyperspectral surface reflectance, and S5P Level-2 CH$_4$ products~\cite{sentinel2l2a,landsatc2l2,emitl2a,s5pch4}. 
Each event is represented as a partial sensor set
\begin{equation}
    \mathcal{X}_i
    =
    \left\{
    (\mathbf{X}_i^s,m_i^s):
    \mathbf{X}_i^s\in
    \mathbb{R}^{T\times C_s\times H_i^s\times W_i^s},
    \ m_i^s\in\{0,1\}
    \right\}_{s\in\mathcal{S}},
    \label{eq:partial-sensor-set}
\end{equation}
where $T$ denotes the temporal context length, $C_s$ is the sensor-specific spectral or product channel dimension, and $H_i^s\times W_i^s$ is the crop size determined by the corresponding training view. 
The binary variable $m_i^s$ indicates whether sensor $s$ is available for event $i$, and the observed sensor subset is
\begin{equation}
    \mathcal{M}_i=\{s\in\mathcal{S}:m_i^s=1\}.
\end{equation}
Only sensors in $\mathcal{M}_i$ are used for tokenization, encoding, fusion, and inference.

For each available sensor, we retrieve a temporal context consisting of an event-day observation near the Carbon Mapper timestamp, a seasonal-history observation around three months before the event, and a long-term-background observation around one year before the event. In the implementation, we set $T=3$ and stack these temporal observations along the channel dimension before tokenization. Thus, the raw input channel dimension for sensor $s$ is $3C_s$: S2 uses 12 surface-reflectance bands, L8/9 uses 7 surface-reflectance bands, EMIT is converted into a compact 16-band methane-relevant spectral representation, and S5P provides coarse Level-2 CH$_4$ context. The resulting channel-stacked inputs have 36 channels for S2, 21 channels for L8/9, and 48 channels for EMIT before sensor-native tokenization.

Specifically, each EMIT observation is projected through WorldView-3 spectral response functions using radiometric bandpass integration, radiometrically aligned by percentile matching, and reprojected to a local UTM coordinate system on the native 60~m EMIT grid~\cite{helmer2007radiometric}. 
This preprocessing reduces the original 285-band dimensionality, balances the number of input channels across sensors, and preserves shortwave-infrared absorption structure near 2.3~$\mu$m that is relevant to methane detection. 

\begin{figure*}[t]
    \centering
    \includegraphics[width=0.95\textwidth]{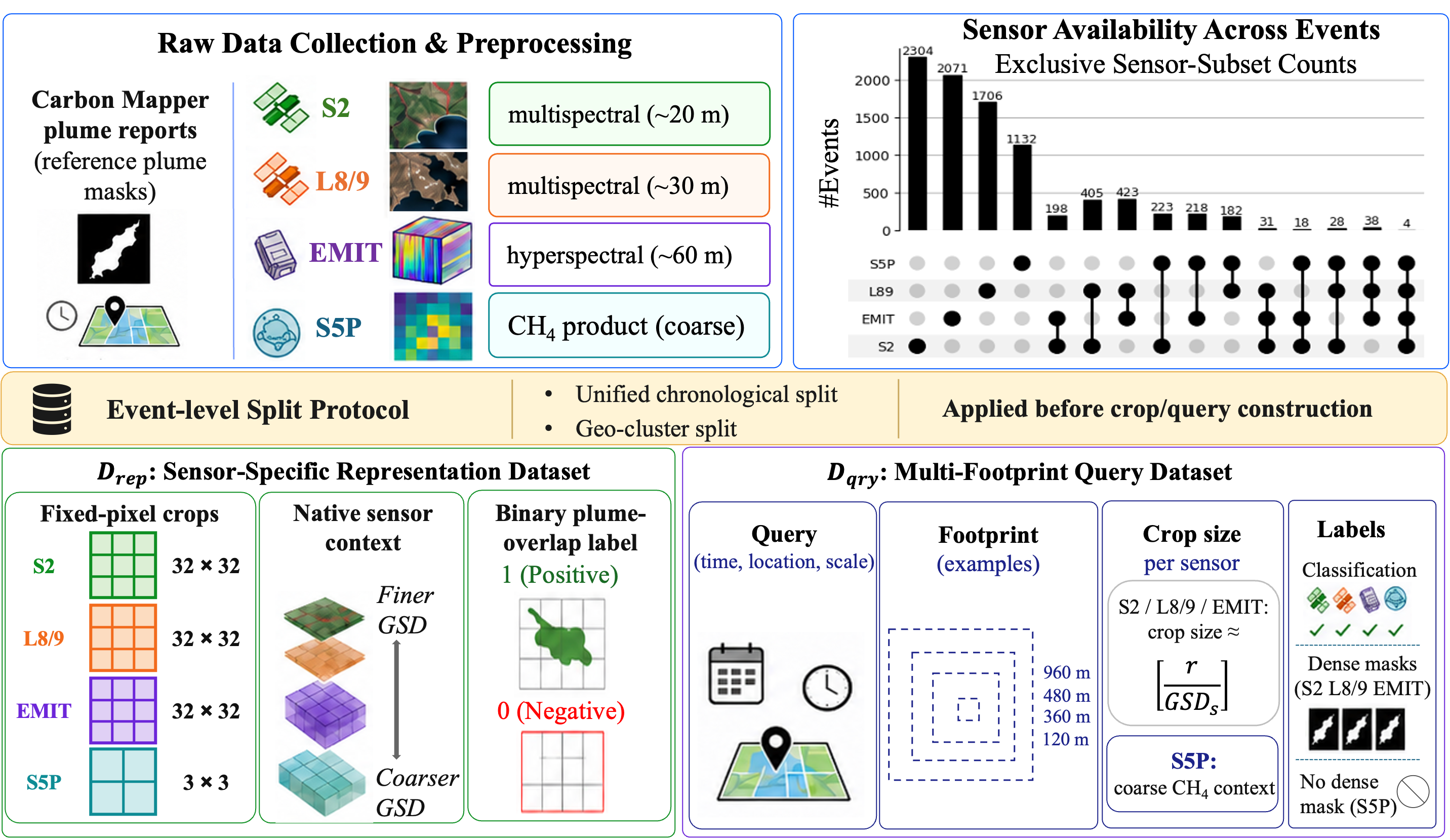}
    \caption{\textit{MethaneUnion} dataset construction pipeline. Carbon Mapper plume reports are used as event anchors, and naturally available
S2, L8/9, EMIT, and S5P observations are matched after sensor-specific preprocessing.
Event-level splitting is applied before crop and query construction, producing two
derived datasets: $\mathcal{D}_{\mathrm{rep}}$ for Stage~1 sensor-native representation
learning and $\mathcal{D}_{\mathrm{qry}}$ for Stage~2 multi-footprint query-level
classification and segmentation.
The sensor-availability panel reports exclusive sensor-subset counts, highlighting
the partial-observation structure of \textit{MethaneUnion}.}
    \label{fig:dataset-benchmark}
\end{figure*}

We partition \textit{MethaneUnion} before constructing Stage~1 crops, Stage~2 queries,
or segmentation masks, so that all observations and derived samples associated
with the same Carbon Mapper plume report remain in the same split. This avoids
evaluating the model on different crops, query footprints, or sensor views
derived from a plume report seen during training. Figure~\ref{fig:dataset-benchmark} summarizes the dataset construction and derived
learning datasets; the chronological and geo-clustered evaluation protocols are
specified in Section~\ref{sec:experiments}.

\subsection{Sensor-Native Representation Learning}

Stage~1 learns methane-aware representations from sensor-native observations before imposing query-level geographic footprints. 
From the split event pool, we construct $\mathcal{D}_{\mathrm{rep}}$ by cropping a fixed number of pixels from each available sensor: $32{\times}32$ pixels for S2, L8/9, and EMIT, and $3{\times}3$ pixels for S5P. 
Because the sensors have different ground sample distances, these fixed-pixel crops correspond to different physical footprints. 
This design preserves the native spatial context exposed by each sensor instead of forcing all sensors into a shared geographic footprint. 
Each crop receives a binary methane label according to whether its geographic footprint overlaps a Carbon Mapper plume mask.

Each available observation $\mathbf{X}_i^s$ is projected into a shared ViT token space while preserving its sensor-native spatial and spectral structure. 
Following a channel-to-patch design, we treat the temporal dimension as additional channel elements, apply a shared per-channel patchifier, inject sensor/channel metadata, and fuse channel tokens within each spatial patch:
\begin{equation}
    \mathbf{Z}_i^s
    =
    \operatorname{Attn}_{\mathrm{ch}}
    \left(
    \operatorname{Patchify}_{\psi}(\mathbf{X}_i^s)
    +
    \mathbf{E}^s
    \right)
    \in
    \mathbb{R}^{N_i^s\times D}.
    \label{eq:token-embedding}
\end{equation}
Here, $N_i^s$ is the number of spatial patches and $D$ is the hidden dimension. 
The metadata embedding $\mathbf{E}^s$ encodes sensor-specific channel semantics, such as wavelength or spectral-response information for optical and hyperspectral sensors, and product/channel identity for derived products such as S5P CH$_4$. 
A CLS token and sensor-specific positional embeddings are then added to form the ViT input sequence $\mathbf{U}_{i,0}^{s}$. 
Thus, heterogeneous sensors remain native before tokenization but share a common transformer interface after tokenization.

For each available sensor $s\in\mathcal{M}_i$, the shared ViT encoder produces a sensor-level CLS representation and patch-token features:
\begin{equation}
    [\mathbf{c}_i^s;\mathbf{P}_i^s]
    =
    \mathcal{F}_{\theta}(\mathbf{U}_{i,0}^{s}),
    \qquad s\in\mathcal{M}_i .
    \label{eq:shared-encoder}
\end{equation}
The encoder is shared across sensors, but each available sensor observation is encoded independently. 
Cross-sensor interaction is performed after encoding by aggregating the observed CLS representations.

To fuse the partially observed sensor set, \textit{MethaneFuse} adds a learned sensor identity embedding $\mathbf{a}^s$ to each CLS representation and applies masked attention pooling (MAP) over the available sensors:
\begin{equation}
    \mathbf{z}_i
    =
    \operatorname{MAP}
    \left(
    \{\mathbf{c}_i^s+\mathbf{a}^s\}_{s\in\mathcal{M}_i}
    \right).
    \label{eq:fused-representation}
\end{equation}
The availability mask determines which sensor tokens enter fusion, allowing the same model to handle arbitrary observed subsets.

The Stage~1 objective combines fused methane supervision with auxiliary sensor-level supervision:
\begin{equation}
    \mathcal{L}_{\mathrm{rep}}
    =
    \mathcal{L}_{\mathrm{fuse}}
    +
    \lambda_{\mathrm{aux}}\mathcal{L}_{\mathrm{aux}} .
    \label{eq:pretrain-objective}
\end{equation}
Here, $\mathcal{L}_{\mathrm{fuse}}$ is the cross-entropy loss from the fused prediction $h_{\mathrm{fuse}}(\mathbf{z}_i)$, and $\mathcal{L}_{\mathrm{aux}}$ is the averaged sensor-level auxiliary cross-entropy over observed CLS representations. 
The auxiliary heads regularize sensor-level representation learning during Stage~1, while the fused head learns to predict from incomplete sensor sets.

\subsection{Sensor-Aware Adaptation for Query-Level Detection}

Stage~2 adapts the Stage~1 representation to query-level methane classification and plume segmentation. 
From the same split event pool, we construct $\mathcal{D}_{\mathrm{qry}}$, where each query is indexed by $q$ and defined by
\begin{equation}
    q = (i,t_q, \ell_q, r_q),
    \label{eq:query-definition}
\end{equation}
where $i$ is the source event index, $t_q$ is the query timestamp, $\ell_q$ is the query center, and $r_q$ specifies the geographic footprint used to construct the query observation. 
Accordingly, Fig.~\ref{fig:model-architecture} writes the query-level partial sensor set as $\{(\mathbf{X}_q^s,m_q^s)\}_{s\in\mathcal{S}}$, where $\mathbf{X}_q^s$ is the sensor-$s$ observation for query $q$ and $m_q^s$ indicates whether that observation is available; the observed query-level sensor subset is $\mathcal{M}_q=\{s\in\mathcal{S}:m_q^s=1\}$. 
For S2, L8/9, and EMIT, the sensor-specific crop size is determined by the corresponding ground sample distance, approximately $\lceil r_q/\mathrm{GSD}(s)\rceil$ pixels per side. 
S5P is retained as a coarse contextual CH$_4$ view because its footprint is much larger than the plume-localization footprints considered in our evaluation.

Each query receives a binary classification label according to whether its geographic footprint overlaps a Carbon Mapper plume mask. 
Dense segmentation masks are generated by reprojecting the Carbon Mapper plume mask to the sensor grid for S2, L8/9, and EMIT; S5P is excluded from dense mask supervision because its coarse footprint does not support local plume delineation. 
In the experiments, we instantiate this query protocol with predefined evaluation footprints to assess detection performance under controlled spatial extents.

\begin{figure}[H]
    \centering
    \includegraphics[width=\columnwidth]{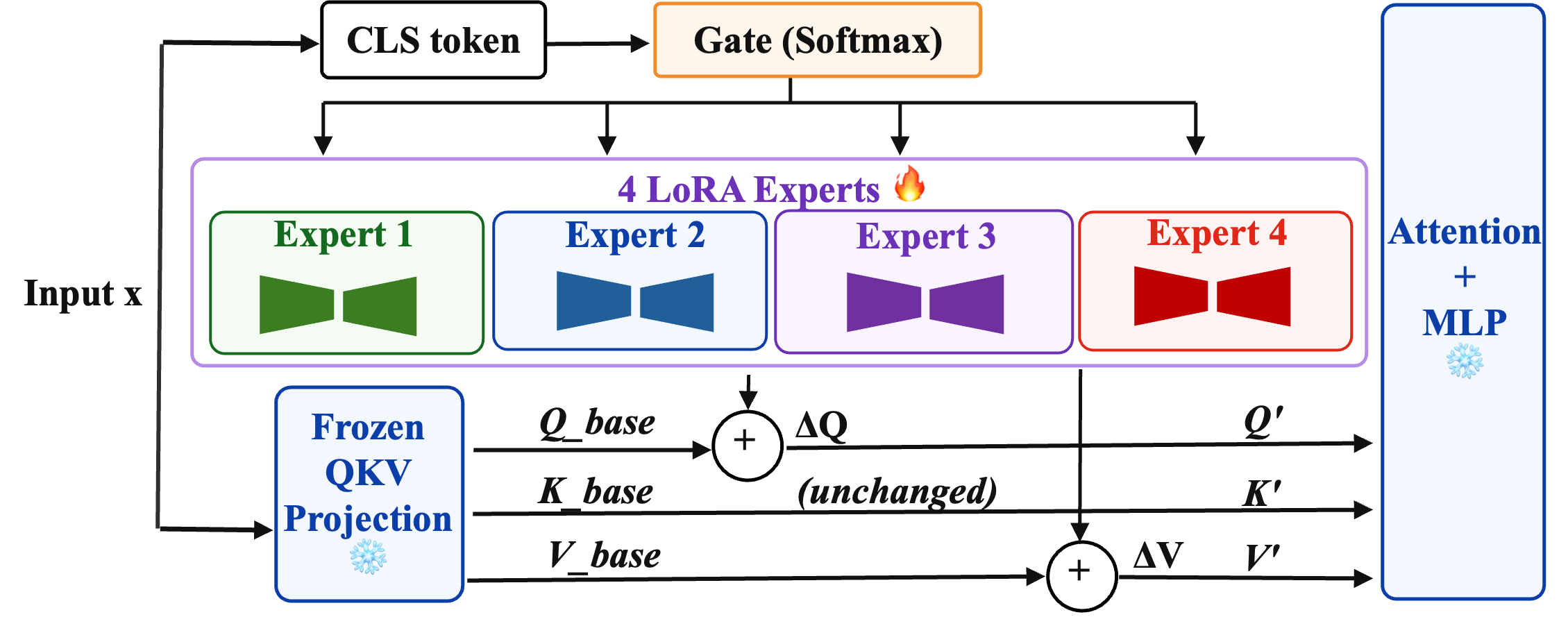}
    \caption{CLS-conditioned LoRA expert adaptation used in Stage~2. The current CLS representation routes each input through a soft mixture of LoRA expert residuals, while the frozen attention projection remains unchanged.}
    \label{fig:loramoe-adapter}
\end{figure}

To adapt the learned representation, Stage~2 freezes the shared weights $\theta^\star$ and inserts trainable LoRA expert adapters into the query and value projections of each ViT block.
Figure~\ref{fig:loramoe-adapter} summarizes the CLS-conditioned routing and expert-residual adaptation used in this stage.

For an input token sequence $\mathbf{X}^{(\ell)}$ at layer $\ell$, the router first extracts the current CLS token
\begin{equation}
    \mathbf{g}^{(\ell)}
    =
    \operatorname{softmax}
    \left(
    \mathbf{W}_{g}^{(\ell)}
    \mathbf{x}_{\mathrm{cls}}^{(\ell)}
    \right),
    \qquad
    \mathbf{x}_{\mathrm{cls}}^{(\ell)}=\mathbf{X}^{(\ell)}_{[:,0,:]},
    \label{eq:cls-routing}
\end{equation}
where $\mathbf{g}^{(\ell)}\in\mathbb{R}^{K}$ gives the mixture weights over $K$ LoRA experts. 
In our implementation, $K$ is set to the number of supported sensors, but the router does not receive an explicit sensor ID. 
Instead, expert selection is conditioned on the current CLS representation, which reflects the sensor-native input, scene content, and intermediate transformer state.

For a frozen attention projection $\mathbf{W}\in\{\mathbf{W}_Q,\mathbf{W}_V\}$, the adapted projection is
\begin{equation}
\begin{aligned}
    \mathbf{W}\mathbf{x}
    &\longrightarrow
    \mathbf{W}\mathbf{x}
    +
    \sum_{k=1}^{K}
    g_k^{(\ell)}
    \Delta_k^{(\ell)}(\mathbf{x}),\\
    \Delta_k^{(\ell)}(\mathbf{x})
    &=
    \frac{\alpha}{r}
    \mathbf{B}_{k}^{(\ell)}
    \mathbf{A}_{k}^{(\ell)}
    \mathbf{x}.
\end{aligned}
    \label{eq:lora-moe}
\end{equation}
where $r$ is the LoRA rank and $\alpha$ is the scaling factor. Here, $\mathbf{A}_{k}^{(\ell)}$ and $\mathbf{B}_{k}^{(\ell)}$ are the down- and up-projection matrices of the $k$-th LoRA expert at layer $\ell$. 
The base projection remains frozen, while the expert residuals provide lightweight input-conditioned adaptation. 
Although the number of experts matches the number of sensors, the experts are not hard-assigned to sensors; they are softly selected by the CLS-token router.

After the adapted encoder, each available sensor produces an adapted CLS representation $\widehat{\mathbf{c}}_q^s$ and adapted patch-token features $\widehat{\mathbf{P}}_q^s$.
For query-level classification, the adapted CLS representations are fused by masked attention pooling.
At this fusion stage, a learned sensor identity embedding is added to each sensor-level CLS feature before attention aggregation:
\begin{equation}
    \widehat{\mathbf{z}}_q
    =
    \operatorname{MAP}
    \left(
    \{\widehat{\mathbf{c}}_q^s+\mathbf{a}^{s}\}_{s\in\mathcal{M}_q}
    \right).
    \label{eq:ada-fused-representation}
\end{equation}
Thus, Stage~2 uses CLS-conditioned expert routing inside the frozen encoder and sensor-embedded masked fusion across available sensors.

For dense segmentation, a lightweight patch-token segmentation head is applied to sensors with meaningful plume-mask supervision, namely S2, L8/9, and EMIT.
S5P is retained only as coarse atmospheric CH$_4$ context for classification because its footprint does not support local plume-boundary delineation.
Segmentation fusion is performed as probability-level late fusion: each available segmentation sensor first predicts an independent plume probability map, the maps are aligned to the Carbon Mapper plume-mask reference grid by geospatial reprojection when metadata is available or by resizing otherwise, and the aligned probabilities are averaged and thresholded at 0.5.
This keeps classification fusion and segmentation fusion separate: classification uses learned masked fusion over adapted CLS representations, while segmentation uses late fusion over aligned per-sensor probability masks.

At inference time, \textit{MethaneFuse} follows the same partial-observation protocol as training.
Given a query anchor, only the naturally available sensors in $\mathcal{M}_q$ are tokenized and encoded.
For classification, the available sensor-level CLS representations are aggregated through learned masked sensor-set fusion.
For segmentation, \textit{MethaneFuse}-Seg produces per-sensor probability maps for available S2, L8/9, and EMIT observations and applies the late-fusion procedure described above.
The predefined query footprints are used as controlled evaluation settings, while the central inference problem remains predicting methane occurrence and plume support from naturally available heterogeneous observations.

\section{Experiments}\label{sec:experiments}
\subsection{Experimental Setup and Baselines}

\textbf{Tasks and labels.}
For query-level classification, samples are labeled by overlap with the Carbon Mapper reference plume mask. Positive queries are those whose geographic footprint overlaps the reference plume mask. Negative queries are sampled from the same event-centered satellite acquisition but at least 5 km away from the reported plume location, and are retained only if their footprint has zero overlap with the reference plume mask at the evaluated scale. This provides local off-plume background samples from the same acquisition context while avoiding ambiguous plume-overlapping negatives.
We use balanced positive and negative sampling, with approximately 50\% positive samples.

\begin{figure}[t]
\centering
\includegraphics[width=\columnwidth]{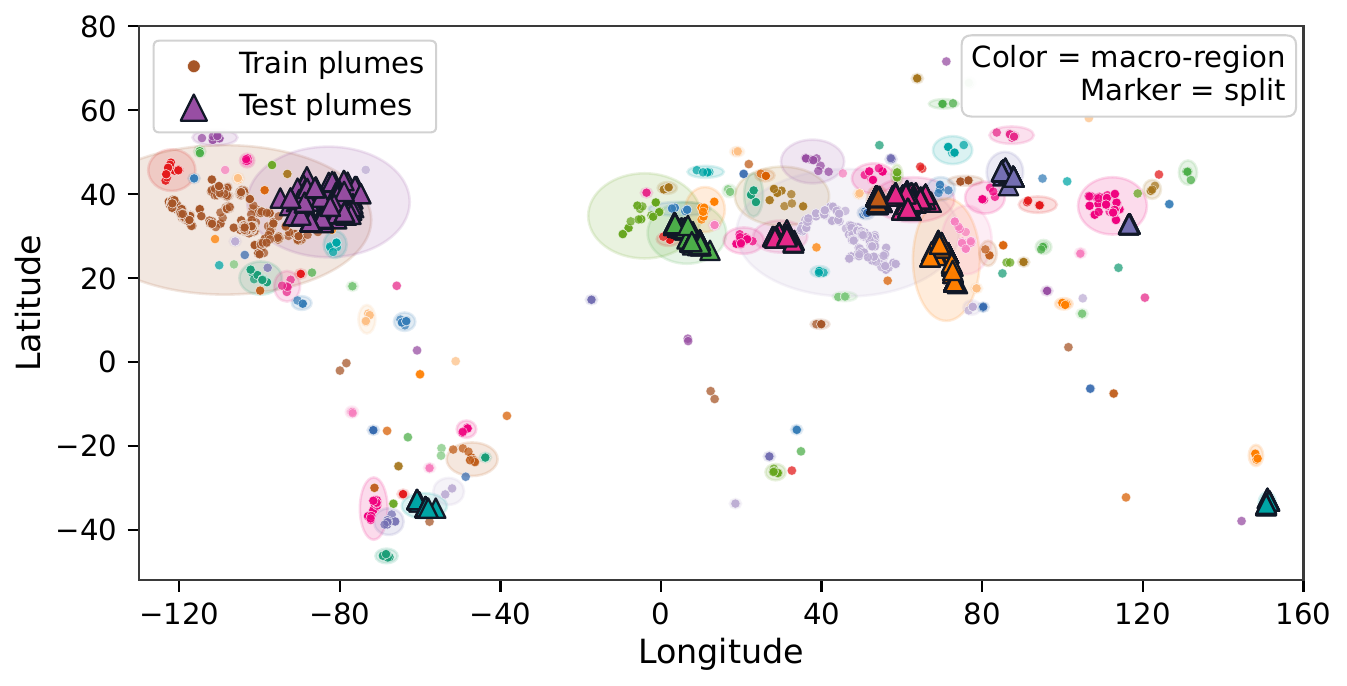}
\caption{Geo-cluster split for spatial robustness evaluation. Plume reports are grouped into tile-scale clusters and 275~km macro-regions, and each macro-region is assigned entirely to training or testing.}
\label{fig:geo-cluster-split}
\end{figure}

\textbf{Evaluation protocols and sample statistics.} 
We use two event-level evaluation protocols. The primary protocol is a chronological split with May 16, 2025 as the cutoff date: Carbon Mapper plume reports on or before the cutoff are assigned to training, and later reports are assigned to testing. The split is applied before constructing sensor-specific crops, scale-controlled queries, and segmentation masks, so all samples derived from the same plume report remain in the same partition. The query-level metadata contains approximately 625k generated samples under the default setting, with about 498k training samples and 126k testing samples, corresponding to an approximately 8/2 partition at the generated-sample level. As a complementary spatial-robustness protocol, we use a geo-clustered split, where plume reports are grouped into spatial macro-regions and each region is assigned entirely to training or testing. The held-out test macro-regions are visualized in Fig.~\ref{fig:geo-cluster-split}.

\textbf{Metrics.}
Classification performance is measured by F1, accuracy, false positive rate (FPR), recall, and AUROC. Segmentation performance is measured by IoU+ following MethaneS2CM~\cite{liu2025methanes2cm}. IoU+ uses standard IoU for non-empty reference plume masks, while assigning 1 to correctly predicted empty masks and 0 to false plume predictions on empty reference masks. We emphasize FPR and AUROC alongside F1 because methane plume classification is strongly affected by sensor-specific background artifacts and false alarms.

\textbf{Baselines.}
We compare \textit{MethaneFuse} with two controlled baseline groups on \textit{MethaneUnion}. First, independent sensor predictors are trained separately for each sensor stream using the same sensor-native tokenization and ViT-style encoder design, and their available prediction scores are combined at inference time using majority voting, average score fusion, or logical OR. Second, we evaluate generic EO representation transfer by fine-tuning representative EO foundation models, including SatMAE, AnySat, and Panopticon, when their input interfaces can be adapted to the corresponding sensor products.

\textbf{Implementation.}
Training was conducted on two NVIDIA A100 80GB GPUs. Stage 1 training
required approximately 44 GPU-hours, and Stage 2 adaptation for the 480~m
setting required approximately 8 GPU-hours.
For Stage~2 adaptation, we used $K=4$ experts, matching the number of
supported sensors, with LoRA rank $r=8$ and scaling factor
$\alpha=16$ for all adapted query and value projections. \textit{MethaneFuse} trains 137.54M parameters in Stage~1 and only 1.22M parameters in Stage~2.
By comparison, the trainable parameter counts are 346.40M for the per-sensor ViT baseline, 342.52M for SatMAE-FT, 503.64M for AnySat-FT, and 395.92M for Panopticon-FT.
These baselines use separate sensor-specific backbones, whereas \textit{MethaneFuse} shares one backbone across sensors and only trains lightweight adapters during downstream adaptation.

\subsection{Main Query-Level Classification Results}

Table~\ref{tab:main-classification} reports the main query-level
classification results under naturally incomplete sensor availability
at the 480~m footprint. Each query is evaluated using its available
sensor subset, and multi-sensor predictions are produced either by
heuristic score-level aggregation or by learned masked sensor-set
fusion. \textit{MethaneFuse} is reported after Stage~2 adaptation, and the adaptation strategy is ablated in Table~\ref{tab:ablation-fusion}.

The heuristic baselines show that score-level fusion can combine
heterogeneous sensor predictions to some extent. Average score fusion gives the
strongest heuristic result with 79.22 F1 and 85.32 AUROC, while Logical
OR increases recall to 80.36 at the cost of a higher 26.96 FPR. Generic
EO representation transfer is weaker in this methane-specific setting:
Panopticon-FT reaches 77.65 F1 and 83.28 AUROC, while SatMAE-FT and
AnySat-FT lag further behind. \textit{MethaneFuse} achieves the strongest
performance across all reported metrics, improving over average score fusion by
+5.65 F1, +6.21 accuracy, +4.46 recall, and +8.30 AUROC while reducing
FPR by 8.19 points.
These results indicate that methane-supervised representation learning
with learned sensor-set fusion is more effective than score-level
fusion or generic EO transfer under partial sensor availability.

\begin{table}[t]
\centering
\caption{Main query-level classification under partial sensor availability at the 480~m footprint. Values are percentages.}
\label{tab:main-classification}
\scriptsize
\setlength{\tabcolsep}{3.1pt}
\begin{tabular}{l l c c c c c}
\toprule
Method & Fusion & F1$\uparrow$ & Acc.$\uparrow$ & FPR$\downarrow$ & Recall$\uparrow$ & AUROC$\uparrow$ \\
\midrule
Per-sensor ViT & Majority & 78.21 & 77.11 & 23.14 & 77.33 & 77.81 \\
Per-sensor ViT & Logical OR & 78.72 & 76.93 & 26.96 & 80.36 & 84.89 \\
Per-sensor ViT & Average & 79.22 & 78.00 & 23.06 & 78.94 & 85.32 \\
\midrule
SatMAE-FT      & Average & 67.60 & 63.33 & 48.41 & 74.44 & 67.61 \\
AnySat-FT      & Average & 58.90 & 58.96 & 37.53 & 55.82 & 62.48 \\
Panopticon-FT  & Average & 77.65 & 75.61 & 29.13 & 79.80 & 83.28 \\
\midrule
\textit{MethaneFuse} & Learned & \textbf{84.87} & \textbf{84.21} & \textbf{14.87} & \textbf{83.40} & \textbf{93.62} \\
\bottomrule
\end{tabular}
\end{table}

\begin{figure}[t]
    \centering
    \includegraphics[width=\columnwidth]{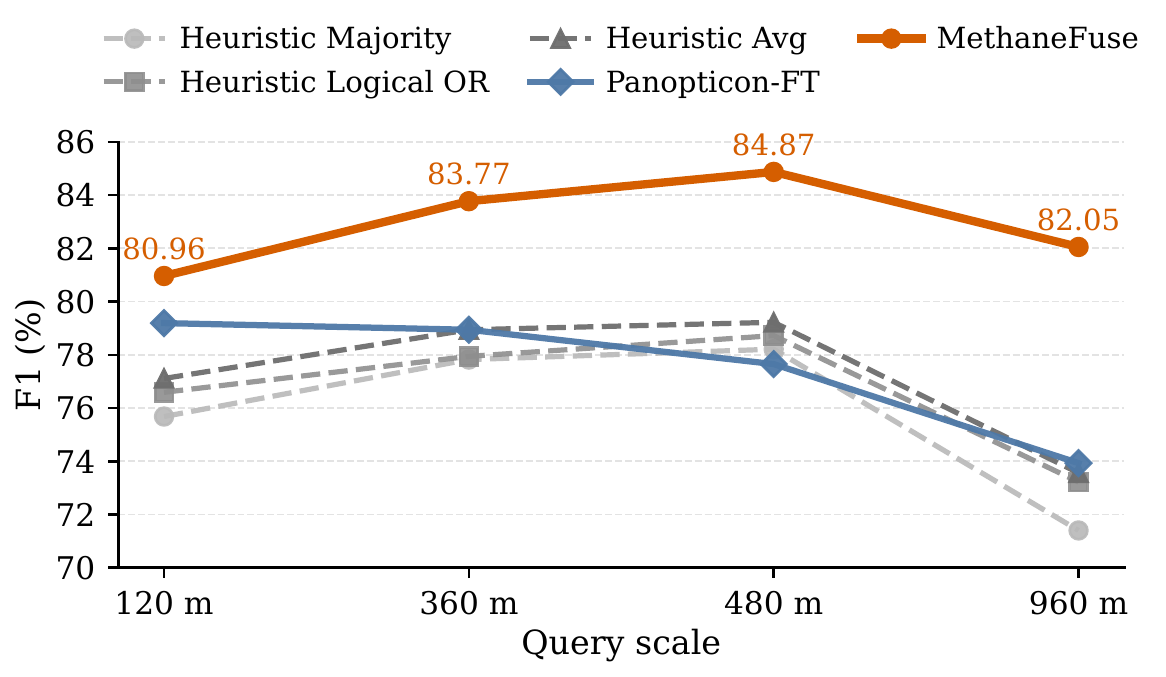}
    \caption{F1 sensitivity across predefined query footprints for \textit{MethaneFuse} and baselines.}
    \label{fig:query-scale-f1-baselines}
\end{figure}

Figure~\ref{fig:query-scale-f1-baselines} summarizes F1 across the four
query footprints. \textit{MethaneFuse} remains above the heuristic fusion
baselines at 120~m, 360~m, 480~m, and 960~m, showing that the learned
partial-observation representation is not tied to a single query scale.

\subsection{Cross-Sensor Knowledge Transfer}

We further test whether \textit{MethaneFuse} transfers methane-relevant knowledge across sensor sets.
For single-sensor regimes, the baseline is an independently trained per-sensor ViT.
For multi-sensor regimes, the baseline averages the independently trained per-sensor ViT scores over the same available sensor group, while \textit{MethaneFuse} uses learned masked sensor-set fusion and Stage 2 adaptation.

\begin{figure}[t]
\centering
\includegraphics[width=\columnwidth]{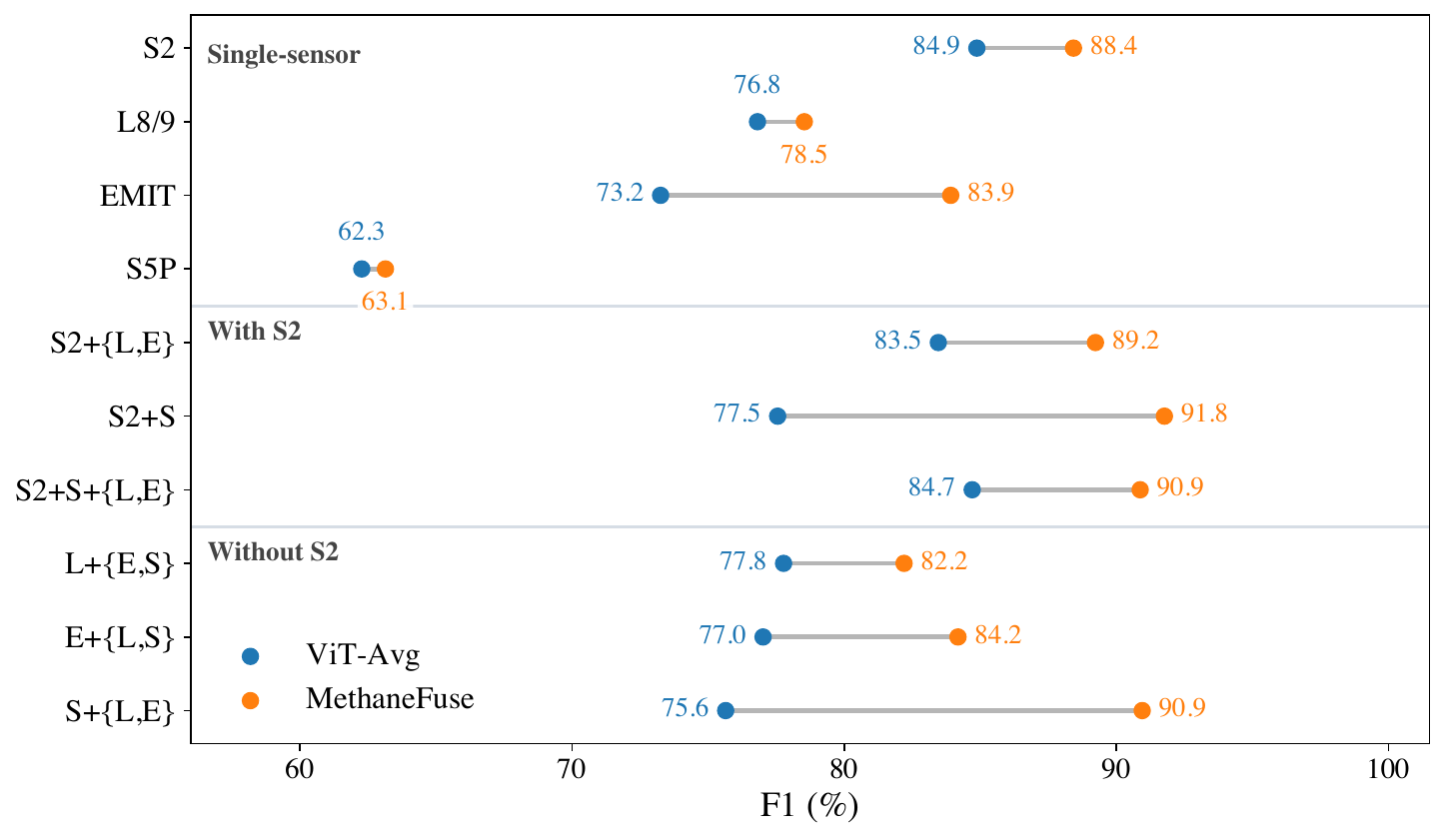}
\caption{Sensor-set transfer at the 480~m footprint measured by F1. Rows are grouped into single-sensor, S2-present, and S2-absent regimes. L, E, and S denote L8/9, EMIT, and S5P.}
\label{fig:sensor-set-f1}
\end{figure}

Figure~\ref{fig:sensor-set-f1} shows that \textit{MethaneFuse} improves F1 across most availability regimes, rather than only in fully multi-sensor settings.
In the single-sensor rows, \textit{MethaneFuse} improves all four streams, with the largest gain on EMIT, where F1 increases from 73.25 to 83.91.
This suggests that the methane-supervised representation learned from irregular multi-sensor observations can also strengthen individual sensor streams.
In the multi-sensor rows, \textit{MethaneFuse} generally improves over average-score fusion, indicating that learned sensor-set fusion is more effective than treating independently trained sensor scores as interchangeable evidence.
The strongest gains appear in several regimes involving weaker or coarser streams, showing that \textit{MethaneFuse} can transfer useful methane-discriminative structure across heterogeneous observation subsets.

\begin{figure}[t]
\centering
\includegraphics[width=\columnwidth]{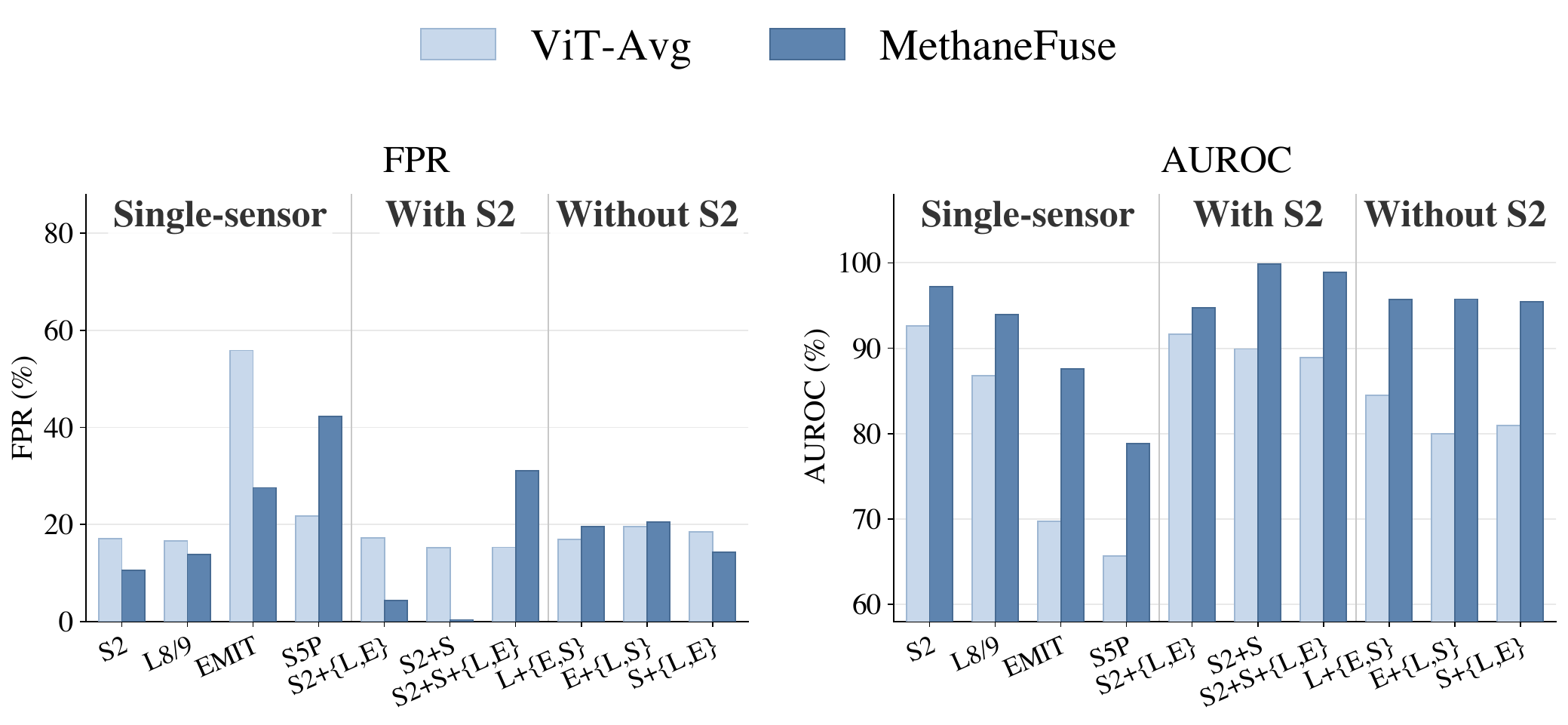}
\caption{FPR and AUROC for the sensor-set transfer groups in Fig.~\ref{fig:sensor-set-f1}. Bars compare the per-sensor ViT average-fusion baseline with \textit{MethaneFuse}.}
\label{fig:sensor-set-fpr-auroc}
\end{figure}

Figure~\ref{fig:sensor-set-fpr-auroc} provides additional insight into false-positive control and ranking quality.
\textit{MethaneFuse} improves AUROC for all availability-conditioned groups, showing more consistent plume/background separation across sensor subsets.
The FPR results are more regime-dependent: \textit{MethaneFuse} substantially reduces false positives for several groups, especially EMIT and multiple S2-containing combinations, but does not reduce FPR uniformly for every availability regime.
Together with the F1 results, these patterns indicate that \textit{MethaneFuse} improves sensor-set transfer mainly by learning a more transferable methane representation and a more effective fusion rule, while still reflecting the noise, resolution, and product characteristics of each sensor group.

\subsection{Geographic Generalization under Geo-Cluster Splits}

The primary task in this paper is event-level methane plume detection under naturally available satellite observations.
Accordingly, the chronological split is the main evaluation protocol, because it reflects the operational setting of applying a detector to future plume events collected after the training period.
In this setting, geographic context is part of the observable satellite evidence: methane-emitting facilities, surface backgrounds, acquisition conditions, and recurring source regions are spatially structured.
A detector trained under a temporal split may therefore learn geography-associated source-context and background cues in addition to plume-related spectral evidence.

We include the geo-cluster split as a complementary spatial robustness test.
This protocol evaluates whether \textit{MethaneFuse} can transfer to held-out macro-regions where facility context, surface materials, climate conditions, and sensor artifacts may differ from those seen during training.
To construct this split, Haversine DBSCAN first groups nearby plume reports into tile-scale clusters and then merges them into 275~km macro-regions.
Each macro-region is assigned entirely to either training or testing, with held-out regions selected to contain approximately 20\% of samples while preserving label balance and geographic diversity.
This setting reduces geographic neighborhood overlap and provides a stricter domain-shift evaluation than the deployment-aligned chronological split.

\begin{figure}[t]
\centering
\includegraphics[width=\columnwidth]{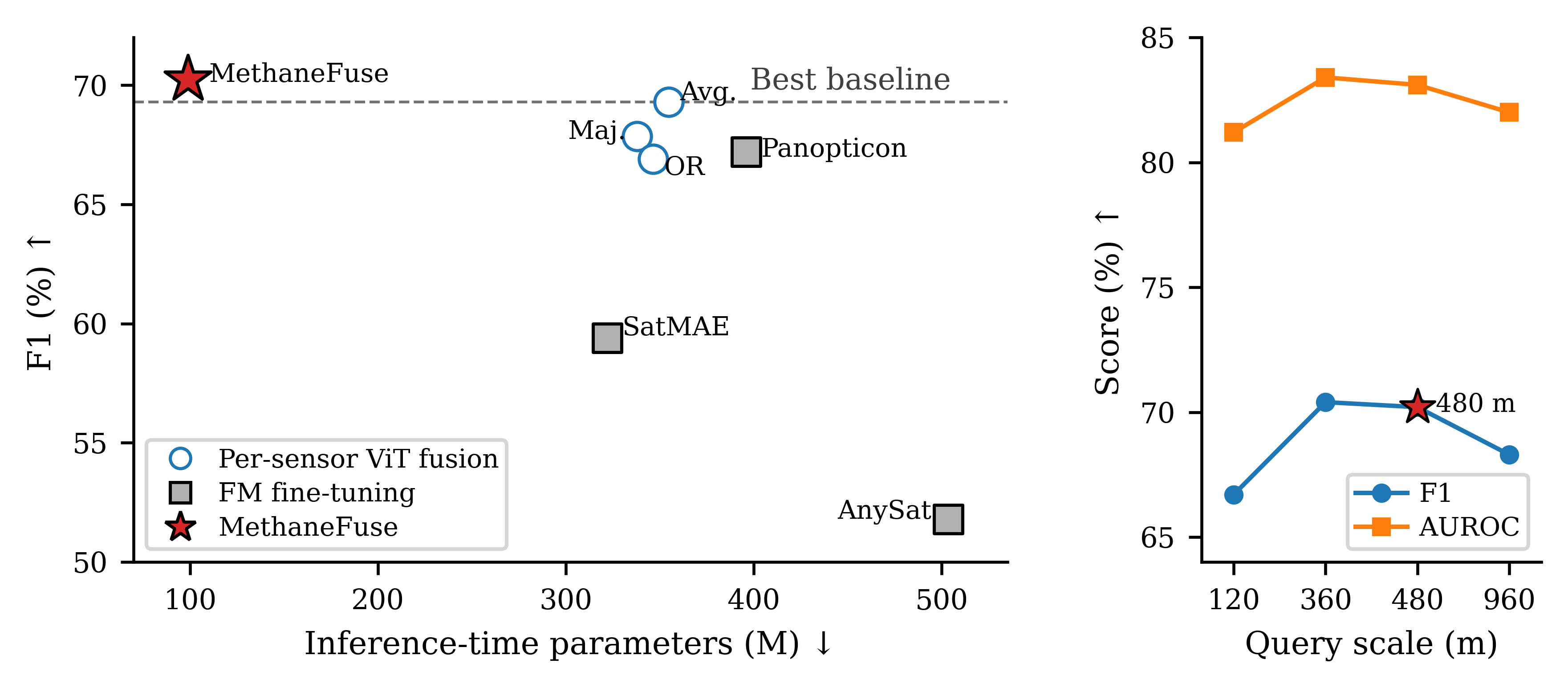}
\caption{Geographic split results under the spatially held-out protocol. The left panel shows F1 versus deployed inference-time parameters at 480~m, and the right panel shows \textit{MethaneFuse} performance across query scales.}
\label{fig:geo-cluster-results}
\end{figure}

Figure~\ref{fig:geo-cluster-results} summarizes the geo-cluster split results.
As expected, performance under spatially held-out macro-regions is lower than under the chronological split, reflecting the removal of region-specific source-context and background information.
At the representative 480 m scale, \textit{MethaneFuse} achieves the highest F1 while using substantially fewer inference-time model parameters than both per-sensor ViT fusion and fine-tuned foundation-model baselines. Here, inference-time parameters count the deployed models used for prediction; for independent per-sensor fusion baselines, the parameters of the separately deployed sensor models are summed.
The scale-wise results further show stable performance across query scales, with the strongest results at intermediate scales of 360--480~m.
These results suggest that \textit{MethaneFuse} retains useful plume-discriminative structure under spatial domain shift, while the chronological split remains the deployment-aligned protocol for the main detection task.

\subsection{Plume Segmentation Results}

Table~\ref{tab:segmentation-results} reports plume segmentation IoU+ (\%) across query scales for S2, L8/9, EMIT, and available-sensor late fusion.
The single-sensor columns evaluate predictions from the corresponding sensor when that sensor is available.
The fusion column evaluates all samples with at least one segmentation-capable sensor among S2, L8/9, and EMIT.
When only one segmentation-capable sensor is available, its predicted probability map is used directly; when multiple sensors are available, their probability maps are aligned to the Carbon Mapper plume-mask reference grid, averaged, and thresholded.
The final prediction is compared with the Carbon Mapper plume mask on the same reference grid using IoU+.
Because plume annotations are derived from satellite plume products with limited spatial precision, temporal mismatch, and sensor-dependent resolution, the segmentation labels should be interpreted as approximate plume support rather than exact plume boundaries.

\begin{table}[t]
\centering
\caption{Plume segmentation IoU+ (\%) across query scales. Fusion uses all available segmentation-capable sensors among S2, L8/9, and EMIT.}
\label{tab:segmentation-results}
\scriptsize
\setlength{\tabcolsep}{4.5pt}
\begin{tabular}{l l c c c c}
\toprule
Scale & Model & S2 & L8/9 & EMIT & Fusion \\
\midrule
\multirow{3}{*}{120~m}
& U-Net & 61.37 & 53.33 & 56.59 & 56.13\\
& Per-sensor ViT & 64.53 & 59.17 & 55.96 & 61.84\\
& \textit{MethaneFuse}-Seg & \textbf{67.69} & \textbf{64.02} & \textbf{56.74} & \textbf{69.99}\\
\midrule
\multirow{3}{*}{360~m}
& U-Net & 52.70 & 46.58 & 56.35 & 56.24\\
& Per-sensor ViT & 61.28 & 56.37 & 56.61 & 57.69 \\
& \textit{MethaneFuse}-Seg & \textbf{68.11} & \textbf{64.04} & \textbf{57.25} & \textbf{60.56}\\
\midrule
\multirow{3}{*}{480~m}
& U-Net & 46.76 & 43.74 & 51.56 & 50.31\\
& Per-sensor ViT & 45.79 & 38.47 & 35.21 & 45.00\\
& \textit{MethaneFuse}-Seg & \textbf{50.16} & \textbf{49.32} & \textbf{55.67} & \textbf{57.49}\\
\midrule
\multirow{3}{*}{960~m}
& U-Net & 41.33 & 41.06 & 50.72 & 34.43\\
& Per-sensor ViT & 26.48 & 35.02 & 49.49 & 37.56\\
& \textit{MethaneFuse}-Seg & \textbf{45.53} & \textbf{43.16} & \textbf{53.10} & \textbf{45.78}\\
\bottomrule
\end{tabular}
\end{table}

\textit{MethaneFuse}-Seg improves IoU+ over the U-Net baseline for most reported sensor-scale pairs, with the strongest gains on S2 and L8/9 at fine and medium query scales.
The improvement on EMIT is more moderate but remains consistent across scales, reflecting the difficulty of plume-boundary localization under coarser and spectrally different observations.
The late-fusion results show that probability-level fusion can improve segmentation when multiple sensor views are available, especially at 120~m and 480~m.
At 360~m and 960~m, fusion is less consistently better than the strongest individual sensor, suggesting that simple mean fusion remains affected by mask alignment uncertainty, resolution mismatch, and imperfect plume annotations.

\subsection{Ablation Study}

We ablate the sensor-adaptive LoRA experts used in Stage~2. The \emph{Full FT} baseline initializes from the Stage~1 encoder and performs full downstream fine-tuning of the encoder and heads. In contrast, \emph{\textit{MethaneFuse}} freezes the Stage~1 shared encoder and updates only the CLS-routed LoRA expert adapters, routing modules, and downstream heads. Table~\ref{tab:ablation-fusion} reports the fusion-level results.

\begin{table}[t]
\centering
\caption{Fusion-level ablation of Stage~2 adaptation strategies. Values are percentages.}
\label{tab:ablation-fusion}
\scriptsize
\setlength{\tabcolsep}{3.1pt}
\begin{tabular}{l l c c c c c}
\toprule
Scale & Variant & F1$\uparrow$ & Acc.$\uparrow$ & FPR$\downarrow$ & Recall$\uparrow$ & AUROC$\uparrow$ \\
\midrule
\multirow{2}{*}{120~m}
& Full FT & 80.23 & 79.32 & 21.10 & 79.70 & \textbf{88.37} \\
& \textit{MethaneFuse} & \textbf{80.96} & \textbf{80.15} & \textbf{19.87} & \textbf{80.17} & 87.44 \\
\midrule
\multirow{2}{*}{360~m}
& Full FT & 82.61 & 81.95 & 17.08 & \textbf{81.09} & \textbf{90.59} \\
& \textit{MethaneFuse} & \textbf{83.77} & \textbf{83.52} & \textbf{13.07} & 80.48 & 90.16 \\
\midrule
\multirow{2}{*}{480~m}
& Full FT & 83.50 & 82.63 & 17.51 & 82.75 & 90.89 \\
& \textit{MethaneFuse} & \textbf{84.87} & \textbf{84.21} & \textbf{14.87} & \textbf{83.40} & \textbf{93.62} \\
\midrule
\multirow{2}{*}{960~m}
& Full FT & 81.03 & 80.29 & 18.74 & \textbf{79.43} & 88.22 \\
& \textit{MethaneFuse} & \textbf{82.05} & \textbf{81.76} & \textbf{14.79} & 78.69 & \textbf{89.11} \\
\bottomrule
\end{tabular}
\end{table}

\textit{MethaneFuse} improves F1, accuracy, and FPR over Full FT at all four query scales. The largest F1 gain occurs at 480~m (+1.37), where \textit{MethaneFuse} also improves AUROC from 90.89 to 93.62. At 120~m and 360~m, Full FT has slightly higher AUROC, but \textit{MethaneFuse} gives better F1 and lower FPR; at 960~m, \textit{MethaneFuse} improves F1 by +1.02 and reduces FPR by 3.95 points. Together, these results support the proposed Stage~2 adaptation strategy as a controlled downstream update for false-positive suppression and scale-controlled classification under heterogeneous partial sensor availability.

\section{Discussion and Conclusion}

This paper studies methane plume detection from incomplete multi-sensor satellite observations, where different plume cases are captured by different subsets of public satellites rather than by complete multi-sensor measurements. We introduced \textbf{\textit{MethaneUnion}}, a temporal multi-sensor dataset built from Carbon Mapper plume reports and matched S2, L8/9, EMIT, and S5P observations. Built on \textit{MethaneUnion}, \textbf{\textit{MethaneFuse}} learns methane plume cues from available sensor subsets through sensor-native representation learning and transfers them to plume classification and segmentation through lightweight sensor-aware adaptation. Across the main classification setting, sensor-availability tests, 120--960~m ground-region evaluations, geo-cluster split, and segmentation experiments, \textit{MethaneFuse} consistently improves over independently trained sensor predictors, heuristic score fusion, and generic EO representation transfer. At the representative 480~m setting, \textit{MethaneFuse} improves over the strongest baseline by 5.65 F1 points and 8.30 AUROC points while reducing false positives by 8.19 points. Sensor-set transfer results further show that \textit{MethaneFuse} strengthens detection when S2 is available and transfers plume knowledge to L8/9, EMIT, and S5P when S2 is unavailable. These findings show that methane plume detection can move beyond S2-only pipelines by learning from incomplete but useful multi-sensor satellite observations.

Several limitations remain. \textit{MethaneUnion} inherits spatial, temporal, and reporting biases from Carbon Mapper plume records and matched satellite acquisitions. Its plume masks provide approximate plume support rather than exact boundaries because of retrieval uncertainty, geolocation error, acquisition-time mismatch, wind-driven displacement, and sensor-dependent resolution. S5P is useful for coarse CH$_4$ classification context but cannot support plume-mask supervision at the evaluated localization scales. Future work will explore uncertainty-aware plume supervision, stronger temporal modeling, improved segmentation fusion, additional satellite products, and deployment-oriented calibration.

\section*{Acknowledgment}
This research received support from the Natural Sciences and Engineering Research Council of Canada (NSERC) Alliance Missions Grant (AMG 577118) and was enabled through collaboration with Enverus. The authors express their gratitude for the data provided by Enverus PRISM.

\end{document}